\pdfoutput=1
\documentclass[11pt]{article}
\usepackage[]{emnlp2021}
\usepackage[T1]{fontenc}
\usepackage{times}
\usepackage{latexsym}

\usepackage{booktabs}
\usepackage{multirow}
\usepackage{amssymb}
\usepackage{graphicx}
\usepackage{arydshln}
\usepackage{enumitem}

\usepackage[table]{colortbl}%
\usepackage{xcolor}

\usepackage{soul}
\usepackage{subcaption}

\usepackage{microtype}

\newcommand\awesome{Awesome-align}
\definecolor{missing}{RGB}{255, 255, 0} %

\definecolor{best}{RGB}{255, 255, 0} %
\newcommand{\best}{\cellcolor{best}}
\definecolor{muchworse}{RGB}{255, 100, 100} %
\newcommand{\muchworse}{\cellcolor{muchworse}}
\definecolor{muchbetter}{RGB}{100, 255, 100} %
\newcommand{\muchbetter}{\cellcolor{muchbetter}}

\definecolor{worse}{RGB}{255, 200, 200} %
\newcommand{\worse}{\cellcolor{worse}}
\definecolor{better}{RGB}{200, 255, 200} %
\newcommand{\better}{\cellcolor{better}}
\newcommand\Mark[1]{\textsuperscript#1}

\title{Everything Is All It Takes: A Multipronged Strategy for Zero-Shot Cross-Lingual Information Extraction}

\author{Mahsa Yarmohammadi\Mark{1}\thanks{~~Equal contribution}, Shijie Wu\Mark{1}\footnotemark[1], Marc Marone\Mark{1}, Haoran Xu\Mark{1}, Seth Ebner\Mark{1}, \\
{\bf Guanghui Qin\Mark{1}},
{\bf Yunmo Chen\Mark{1}},
{\bf Jialiang Guo\Mark{1}},
{\bf Craig Harman\Mark{1}}, 
{\bf Kenton Murray\Mark{1}}, \\
{\bf Aaron Steven White\Mark{2}},
{\bf Mark Dredze\Mark{1}}, 
{\bf Benjamin Van Durme\Mark{1}}\\[1em]
\Mark{1}Johns Hopkins University, \Mark{2}University of Rochester\\[1em]
\texttt{\{mahsa,shijie.wu,vandurme\}@jhu.edu}\\[1em]
}
\date{}

\begin{document}
\maketitle
\begin{abstract}

Zero-shot cross-lingual information extraction (IE) describes the construction of an IE model for some target language, given existing annotations exclusively in some other language, typically English.  While the advance of pretrained multilingual encoders suggests an easy optimism of "train on English, run on any language", we find through a thorough exploration and extension of techniques that a combination of approaches, both new and old, leads to better performance than any one cross-lingual strategy in particular. We explore techniques including data projection and self-training, and how different pretrained encoders impact them. We use English-to-Arabic IE as our initial example, demonstrating strong performance in this setting for event extraction, named entity recognition, part-of-speech tagging, and dependency parsing.
We then apply data projection and self-training to three tasks across eight target languages. Because no single set of techniques performs the best across all tasks, we encourage practitioners to explore various configurations of the techniques described in this work when seeking to improve on zero-shot training.
\end{abstract}

\section{Introduction}

We consider zero-shot cross-lingual information extraction (IE), in which training data exists in a source language but not in a target language.  Massively multilingual encoders like Multilingual BERT \citep[mBERT;][]{devlin-etal-2019-bert} and XLM-RoBERTa \citep[XLM-R;][]{conneau-etal-2020-unsupervised} allow for a strategy of training only on the source language data, trusting entirely in a shared underlying feature representation across languages~\cite{wu-dredze-2019-beto,conneau-etal-2020-emerging}. However, in meta-benchmarks like XTREME \cite{hu2020xtreme}, such cross-lingual performance on structured prediction tasks is far behind that on sentence-level or retrieval tasks \cite{Ruder2021XTREMERTM}: performance in the target language is often far below that of the source language.  Before multilingual encoders, cross-lingual IE was approached largely as a data projection problem: one either translated and aligned the source training data to the target language, or at test time one translated target language inputs to the source language for prediction \cite{yarowsky-ngai-2001-inducing}.

We show that by augmenting the source language training data with data in the target language---either via projection of the source data to the target language (so-called ``silver'' data) or via self-training with translated text---zero-shot performance can be improved. Further improvements might come from using better pretrained encoders or improving on a projection strategy through better automatic translation models or better alignment models. In this paper, we explore all the options above, finding that \emph{everything is all it takes} to achieve our best experimental results, suggesting that a silver bullet strategy does not currently exist.

Specifically, we evaluate: cross-lingual data projection techniques with different machine translation and word alignment components, the impact of bilingual and multilingual contextualized encoders on each data projection component, and the use of different encoders in task-specific models. We also offer suggestions for practitioners operating under different computation budgets on four tasks: event extraction, named entity recognition, part-of-speech tagging, and dependency parsing, following recent work that uses English-to-Arabic tasks as a test bed \cite{lan-etal-2020-empirical}.
We then apply data projection and self-training to three structured prediction tasks---named entity recognition, part-of-speech tagging, and dependency parsing---in multiple target languages. Additionally, we use self-training as a control against data projection to determine in which situations data projection improves performance.

\begin{figure*}[t]
    \centering
    \includegraphics[width=1\textwidth]{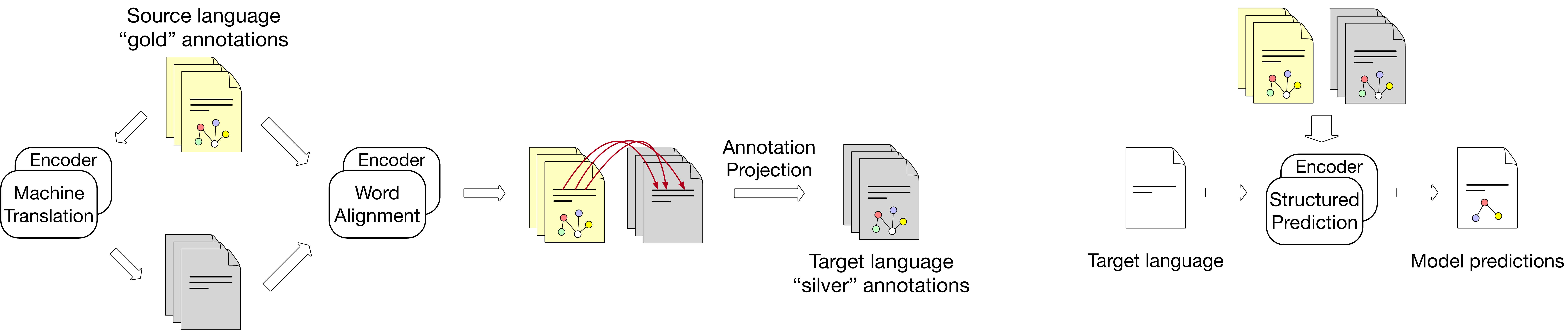}
    \caption{Process for creating projected ``silver'' data from source ``gold'' data (left). Downstream models are trained on a combination of gold and silver data (right). Components in boxes have learned parameters.}
    \label{fig:system}
\end{figure*}

Our contributions include the following:

\begin{itemize}[itemsep=-1pt]
    \item examination of the impact of statistical and neural word aligners and publicly available and custom machine translation (MT) models on annotation projection, 
    \item examination of the impact of publicly available and custom multilingual and bilingual encoders of different model sizes, both as the basis of models for downstream tasks and as components of word aligners and MT models, %
    \item use of self-training on translated text as a way to automatically create labeled target language data and as a controlled comparison to analyze when data projection helps, and 
    \item extensive experiments demonstrating improvements over zero-shot transfer and analysis showing that the best setup is task dependent.
\end{itemize}
We also make available models and tools that enabled our analysis.

\section{Universal Encoders}
\label{sec:encoder}

While massively multilingual encoders like mBERT and XLM-R enable strong zero-shot cross-lingual performance \cite{wu-dredze-2019-beto,conneau-etal-2020-unsupervised}, they suffer from the curse of multilinguality \cite{conneau-etal-2020-unsupervised}: cross-lingual effectiveness suffers as the number of supported languages increases for a fixed model size.
We would therefore expect that when restricted to only the source and target languages, a bilingual model should perform better than (or at least on par with) a multilingual model of the same size, assuming both languages have sufficient corpora \cite{wu-dredze-2020-languages}.
If a practitioner is interested in only a small subset of the supported languages, \textit{is the multilingual model still the best option}?

To answer this question, we use English and Arabic as a test bed. In \autoref{tab:encoder}, we summarize existing publicly available encoders that support both English and Arabic.\footnote{We do not include multilingual T5 \cite{xue-etal-2021-mt5} as it is still an open question on how to best utilize text-to-text models for structured prediction tasks \cite{Ruder2021XTREMERTM}.}
Base models are 12-layer Transformers ($\texttt{d\_model} = 768$), and large models are 24-layer Transformers ($\texttt{d\_model} = 1024$) \cite{vaswani2017attention}.
As there is no publicly available large English--Arabic bilingual encoder, we train two encoders from scratch, named L64K and L128K, with vocabulary sizes of 64K and 128K, respectively.\footnote{L128K available at \url{https://huggingface.co/jhu-clsp/roberta-large-eng-ara-128k}} With these encoders, we can determine the impacts of model size and the number of supported languages.

\section{Data Projection}

We create silver versions of the data by automatically projecting annotations from source English gold data to their corresponding machine translations in the target language.\footnote{Code available at \url{https://github.com/shijie-wu/crosslingual-nlp}}
Data projection transfers word-level annotations in a source language to a target language via word-to-word alignments \cite{yarowsky-etal-2001-inducing}. The technique has been used to create cross-lingual datasets for a variety of structured natural language processing tasks, including named entity recognition \cite{stengel-eskin-etal-2019-discriminative} and semantic role labeling \cite{akbik-etal-2015-generating,aminian-etal-2017-transferring,fei-etal-2020-cross}.

To create silver data, as shown in \autoref{fig:system}, we: (1) translate the source text to the target language using the MT system described in Section \ref{sec:mt}, (2) obtain word alignments between the original and translated parallel text using a word alignment tool, and (3) project the annotations along the word alignments. We then combine silver target data with gold source data to augment the training set for the structured prediction task.

For step (1), we rely on a variety of source-to-target MT systems.
To potentially leverage monolingual data, as well as contextualized cross-lingual information from pretrained encoders, we feed the outputs of the final layer of frozen pretrained encoders as the inputs to the MT encoders. We consider machine translation systems: (i) whose parameters are randomly initialized, (ii) that incorporate information from massively multilingual encoders, and (iii) that incorporate information from bilingual encoders that have been trained on only the source and target languages.

\begin{table}
\small
\begin{center}

\begin{tabular}{l|ll}

\toprule
& \textbf{Base} & \textbf{Large}\\
\midrule
\textbf{Multilingual} & mBERT & XLM-R \\
 & (\citeauthor{devlin-etal-2019-bert}) & (\citeauthor{conneau-etal-2020-unsupervised}) \\
\textbf{Bilingual} & GBv4 & L64K \& L128K \\
 & (\citeauthor{lan-etal-2020-empirical}) &  \textbf{(Ours)} \\
\bottomrule

\end{tabular}
\caption{Encoders supporting English and Arabic.}
\label{tab:encoder}
\end{center}
\end{table}

After translating source sentences to the target language, in step (2) we obtain a mapping of the source words to the target words using publicly available automatic word alignment tools. Similarly to our MT systems, we incorporate contextual encoders in the word aligner. We hypothesize that better word alignment yields better silver data, and better information extraction consequently.

For step (3), we apply direct projection to transfer labels from source sentences to target sentences according to the word alignments. Each target token receives the label of the source token aligned to it (token-based projection).  For multi-token spans, the target span is a contiguous span containing all aligned tokens from the same source span (span-based projection), potentially including tokens not aligned to the source span in the middle. Three of the IE tasks we consider---ACE, named entity recognition, and BETTER---use span-based projection, and we filter out projected target spans that are five times longer than the source spans. Two syntactic tasks---POS tagging and dependency parsing---use token-based projection. For dependency parsing, following \citet{tiedemann-etal-2014-treebank}, we adapt the disambiguation of many-to-one mappings by choosing as the head the node that is highest up in the dependency tree. In the case of a non-aligned dependency head, we choose the closest aligned ancestor as the head.

To address issues like translation shift, filtered projection \cite{akbik-etal-2015-generating,aminian-etal-2017-transferring} has been proposed to obtain higher precision but lower recall projected data. To maintain the same amount of silver data as gold data, in this study we do not use any task-specific filtered projection methods to remove any sentence.

\section{Tasks}
We employ our silver dataset creation approach on a variety of tasks.\footnote{See \autoref{app:ft-hparams} for dataset statistics and fine-tuning hyperparameters for each task.} For English--Arabic experiments, we consider ACE, BETTER, NER, POS tagging, and dependency parsing. For multilingual experiments, we consider NER, POS tagging, and dependency parsing. We use English as the source language and 8 typologically diverse target languages: Arabic, German, Spanish, French, Hindi, Russian, Vietnamese, and Chinese. Because of the high variance of cross-lingual transfer \cite{wu-dredze-2020-explicit}, we report the average test performance of three runs with different predefined random seeds (except for ACE).\footnote{We report one run for ACE due to long fine-tuning time.} For model selection and development, we use the English dev set in the zero-shot scenario and the combined English dev and silver dev sets in the silver data scenario.

\subsection{ACE}
Automatic Content Extraction (ACE) 2005 \cite{ace2005} provides named entity, relation, and event annotations for English, Chinese, and Arabic. We conduct experiments on English as the source language and Arabic as the target language. We use the OneIE framework \cite{lin-etal-2020-joint}, a joint neural model for information extraction, which has shown state-of-the-art results on all subtasks. We use the same hyperparameters as in \citet{lin-etal-2020-joint} for all of our experiments.
We use the OneIE scoring tool to evaluate the prediction of entities, relations, event triggers, event arguments, and argument roles. For English, we use the same English document splits as \cite{lin-etal-2020-joint}. That work does not consider Arabic, so for Arabic we use the document splits from \cite{lan-etal-2020-empirical}.

\subsection{Named Entity Recognition}
We use WikiAnn \citep{pan-etal-2017-cross} for English--Arabic and multilingual experiments. The labeling scheme is BIO with 3 types of named entities: PER, LOC, and ORG. On top of the encoder, we use a linear classification layer with softmax to obtain word-level predictions. The labeling is word-level while the encoders operate at subword-level, thus, we mask the prediction of all subwords except for the first one. We evaluate NER performance by F1 score of the predicted entity.

\subsection{Part-of-speech Tagging}
We use the Universal Dependencies (UD) Treebank \citep[v2.7;][]{zeman-universal-2.7}.\footnote{We use the following treebanks: Arabic-PADT, German-GSD, English-EWT, Spanish-GSD, French-GSD, Hindi-HDTB, Russian-GSD, Vietnamese-VTB, and Chinese-GSD.}
Similar to NER, we use a word-level linear classifier on top of the encoder, and evaluate performance by the accuracy of predicted POS tags.

\subsection{Dependency Parsing}
We use the same treebanks as the POS tagging task. For the task-specific layer, we use the graph-based parser of \citet{DBLP:journals/corr/DozatM16}, but replace their LSTM encoder with our encoders of interest. We follow the same policy as that in NER for masking non-first subwords. We predict only the universal dependency labels, and we evaluate performance by labeled attachment score (LAS). 

\subsection{BETTER}
 The Better Extraction from Text Towards Enhanced Retrieval (BETTER) Program\footnote{\url{https://www.iarpa.gov/index.php/research-programs/better}} develops methods for extracting increasingly fine-grained semantic information in a target language, given gold annotations only in English. 
 We focus on the coarsest ``Abstract'' level, where the goal is 
to identify events and their agents and patients. %
The documents come from the news-specific
portion of Common Crawl.
We report the program-defined ``combined F1'' metric, which is the product of ``event match F1'' and ``argument match F1'', which are based on an alignment of predicted and reference event structures.%

To find all events in a sentence and their corresponding arguments,
we model the structure of the events as a tree, where event triggers are children of the ``virtual root'' of the sentence and arguments are children of event triggers
\citep{cai-etal-2018-full}.
Each node is associated with a span in the text and is labeled with an event or argument type label.

We use a model for event structure prediction that has three major components: a contextualized encoder, tagger, and typer~\cite{xia-etal-2021-lome}.\footnote{Code available at \url{https://github.com/hiaoxui/span-finder}}
The tagger is a BiLSTM-CRF BIO tagger \citep{panchendrarajan-amaresan-2018-bidirectional} trained to predict child spans conditioned on parent spans and labels.
The typer is a feedforward network whose inputs are a parent span representation, parent label embedding, and child span representation.
The tree is produced level-wise at inference time, first predicting event triggers, typing them, and then predicting arguments conditioned on the typed triggers.

\section{Experiments}

\subsection{Universal Encoders}

We train two English--Arabic bilingual encoders.\footnote{Details of pretraining can be found in \autoref{app:pretrain-hparams}.} Both of them are 24-layer Transformers ($\texttt{d\_model} = 1024$), the same size as XLM-R large. We use the same Common Crawl corpus as XLM-R for pretraining. Additionally, we also use English and Arabic Wikipedia, Arabic Gigaword \cite{parker2011arabic}, Arabic OSCAR \cite{ortiz-suarez-etal-2020-monolingual}, Arabic News Corpus \cite{el20161}, and Arabic OSIAN \cite{zeroual-etal-2019-osian}. In total, we train with 9.2B words of Arabic text and 26.8B words of English text, more than either XLM-R (2.9B words/23.6B words) or GBv4 (4.3B words/6.1B words).\footnote{We measure word count with \texttt{wc -w}.} We build two English--Arabic joint vocabularies using SentencePiece \cite{kudo-richardson-2018-sentencepiece}, resulting in two encoders: \textbf{L64K} and \textbf{L128K}. For the latter, we additionally enforce coverage of all Arabic characters after normalization.

\subsection{Machine Translation}
\label{sec:mt}
For all of our MT experiments, we use a dataset of 2M sentences from publicly available data including the UN corpus, Global Voices, wikimatrix, and newscommentary11 \citep{ziemski-etal-2016-united, prokopidis-etal-2016-parallel, schwenk-etal-2021-wikimatrix, callison-burch-etal-2011-findings}. We pre-filtered the data using LASER scores to ensure high quality translations are used for our bitext \citep{schwenk-douze-2017-learning, thompson-post-2020-automatic}.

All of our systems are based on the Transformer architecture \citep{vaswani2017attention}.\footnote{See \autoref{app:mt-hparams} for a full list of hyperparameters.} Our baseline system uses a joint English--Arabic vocabulary with 32k BPE operations \cite{sennrich-etal-2016-neural}. The public system is a publicly released model that has been demonstrated to perform well \citep{tiedemann-2020-tatoeba}.\footnote{The public MT model is available at \url{https://huggingface.co/Helsinki-NLP/opus-mt-en-ar}} The other systems use contextualized embeddings from frozen pretrained language models as inputs to the encoder. For the decoder vocabulary, these systems all use the GBv4 vocabulary regardless of which pretrained language model was used to augment the encoder.

\paragraph{Incorporating Pretrained LMs}
In order to make use of the pretrained language models, we use the output of the last layer of the encoder. A traditional NMT system uses a prespecified, fixed size vocabulary with randomly initialized parameters for the source embedding layer. To incorporate a pretrained language model, we instead use the exact vocabulary of that model. A sentence is fed into the encoder and the resultant vectors from the output layer are used instead of the randomly initialized embedding layer. We freeze these pretrained language models so that no gradient updates are applied to them during MT training, whereas the randomly initialized baselines are updated. A preliminary experiment in \citet{Zhu2020Incorporating} uses a related system that leverages the last layer of BERT. However, that experiment was monolingual, and our hypothesis is that the shared embedding space of a multilingual encoder will aid in training a translation system.

\paragraph{Denormalization System}
Generating text in Arabic is a notoriously difficult problem due to data sparsity problems arising from the morphological richness of the language, frequently necessitating destructive normalization schemes during training that must be heuristically undone in post-processing to ensure well-formed text \citep{sajjad2013qcri}. %
All of the most common multilingual pretrained encoders use a form of destructive normalization which removes diacritics, which causes MT systems to translate into normalized Arabic text. %
To generate valid Arabic text, we train a sequence-to-sequence model that transduces normalized text into unnormalized text using the Arabic side of our bitext, before and after normalization. %
Our transducer uses the same architecture and hyperparameters as our baseline MT system, but with 1k BPE operations instead of 32k. On an internal held-out test set, we get a BLEU score of 96.9 with a unigram score of 98.6, implying few errors will propagate due to the denormalization process.\footnote{Denormalization code available at \url{https://github.com/KentonMurray/ArabicDetokenizer}} %

\paragraph{Intrinsic Evaluation} \autoref{tab:mt} shows the denormalized and detokenized BLEU scores for English--Arabic MT systems with %
different encoders on the IWLST'17 test set using sacreBLEU \citep{post-2018-call}. %
The use of contextualized embeddings from pretrained encoders results in better performance than using a standard randomly initialized MT model regardless of which encoder is used. The best performing system uses our bilingual L64K encoder, but all pretrained encoder-based systems perform well and within 0.5 BLEU points of each other. We hypothesize that the MT systems are able to leverage the shared embedding spaces of the pretrained language models in order to assist with translation.

\begin{table}[]
\small
\begin{center}
\begin{tabular}{cc}
\toprule

\textbf{Encoder} & \textbf{BLEU}  \\
\midrule
Public & 12.7 \\
\midrule
None & 14.9 \\
\midrule
mBERT & 15.7 \\
GBv4 & 15.7 \\
\midrule
XLM-R & 16.0 \\
L64K & \bf 16.2 \\
L128K & 15.8\\

\bottomrule
\end{tabular}
\caption{BLEU scores of MT systems with different pre-trained encoders on English--Arabic IWSLT'17. %
}
\label{tab:mt}
\end{center}
\end{table}

\subsection{Word Alignment}
Until recently, alignments have typically been obtained using unsupervised statistical models such as  GIZA++ \cite{och-ney-2003-systematic} and fast-align \cite{dyer-etal-2013-simple}. Recent work has focused on using the similarities between contextualized embeddings to obtain alignments \cite{jalili-sabet-etal-2020-simalign,daza-frank-2020-x,dou-neubig-2021-word}, achieving state-of-the-art performance.

\begin{table}
\small
\begin{center}
\begin{tabular}{lccccc}

\toprule
\textbf{Model} & \textbf{Layer}\dag & \textbf{AER} & \textbf{P} & \textbf{R}  & \textbf{F}\\
\midrule
fast-align*& n/a & 47.4&53.9&51.4&52.6\\
\midrule
\multicolumn{6}{l}{\textit{Awesome-align w/o FT}} \\
\midrule
mBERT           & 8  & 35.6 & 78.5 & 54.5 & 64.4 \\
GBv4     & 8  & \textbf{32.7} & \textbf{85.6} & 55.4 & \textbf{67.3} \\
\midrule
XLM-R & 16 & 40.1 & 78.6 & 48.4 & 59.9 \\
L64K  & 17 & 34.0 & 81.5 & \textbf{55.5} & 66.0 \\
L128K & 17 & 35.1 & 80.0 & 54.5 & 64.9 \\
\midrule
\multicolumn{6}{l}{\textit{Awesome-align w/ FT}} \\
\midrule
mBERT$_\textit{ft}$   & 8  & 30.0 & 81.9 & \textbf{61.2} & 70.0 \\
GBv4$_\textit{ft}$       & 8  & 29.3 & 86.9 & 59.7 & 70.7 \\
\midrule
XLM-R$_\textit{ft}$   & 18 & \textbf{27.8} & \textbf{90.3} & 60.2 & \textbf{72.2} \\
L64K$_\textit{ft}$    & 17 & 29.1 & 84.9 & 60.9 & 70.9 \\
L128K$_\textit{ft}$   & 16 & 32.2 & 80.3 & 58.7 & 67.8 \\
\midrule
\multicolumn{6}{l}{\textit{Awesome-align w/ FT \& supervision}} \\
\midrule
XLM-R$_\textit{ft.s}$   & 16 & \textbf{23.3} & 92.5 & \textbf{65.6} & \textbf{76.7} \\
L128K$_\textit{ft.s}$   & 17 & 23.5 & \textbf{93.7}& 64.6 & 76.5\\
\bottomrule

\end{tabular}
\caption{Alignment performance on GALE EN--AR. *Trained on MT bitext. \dag We report the best layer of each encoder based on dev alignment error rate (AER).}
\label{tab:alignment-results}
\end{center}
\end{table}

We use two automatic word alignment tools: fast-align, a widely used statistical alignment tool based on IBM models \cite{brown-etal-1993-mathematics}; and \awesome~\cite{dou-neubig-2021-word}, a contextualized embedding-based word aligner that extracts word alignments based on similarities of the tokens' contextualized embeddings. \awesome~achieves state-of-the-art performance on five language pairs. Optionally, \awesome~can be fine-tuned on
parallel text with objectives suitable for word alignment and on gold alignment data.

We benchmark the word aligners on the gold standard alignments in the GALE Arabic--English Parallel Aligned Treebank \cite{li-etal-2012-parallel}. We use the same data splits as \citet{stengel-eskin-etal-2019-discriminative}, containing 1687, 299, and 315 sentence pairs in the train, dev, and test splits, respectively. %
To obtain alignments using fast-align, we append the test data to the MT training bitext and run the tool from scratch.
\awesome~extracts the alignments for the test set based on pretrained contextualized embeddings. These encoders can be fine-tuned using the parallel text in the train and dev sets. Additionally, the encoders can be further fine-tuned using supervision from gold word alignments.

\paragraph{Intrinsic Evaluation} \autoref{tab:alignment-results} shows the performance of word alignment methods on the GALE English--Arabic alignment dataset. \awesome~outperforms fast-align, and fine-tuned \awesome~($\textit{ft}$) outperforms models that were not fine-tuned. Incorporating supervision from the gold alignments ($\textit{s}$) leads to the best performance. %

\section{Cross-lingual Transfer Results}
\label{sec:results}

One might optimistically consider that the latest multilingual encoder (in this case XLM-R) in the zero-shot setting would achieve the best possible performance. However, in our extensive experiments in \autoref{tab:main} and \autoref{tab:multi}, we find that the zero-shot approach can usually be improved with data projection. In this section, we explore the impact of each factor within the data projection process.

\subsection{English--Arabic Experiments}
\label{sec:results-enar}

\begin{table*}[h]
\small
\begin{center}
\begin{tabular}{lll| ccccc|| ccccc}

\toprule
 & \textbf{MT} & \textbf{Align} & \textbf{ACE} & \textbf{NER} & \textbf{POS} & \textbf{Parsing} & \textbf{BET.} & \textbf{ACE} & \textbf{NER} & \textbf{POS} & \textbf{Parsing} & \textbf{BET.}\\
\midrule
& & & \multicolumn{5}{l||}{\textit{mBERT (base, multilingual)}} & \multicolumn{5}{l}{\textit{XLM-R (large, multilingual)}}   \\
\midrule
(Z) & - & - & 27.0 & 41.6 & 59.7 & 29.2 & 39.9  & \textbf{45.1} & 46.4 & 73.3 & \best \textbf{48.0} & 50.8\\
\midrule
(A) & public & FA & \better +2.5 & \worse -3.8 & \muchbetter +8.5 & \muchbetter +7.3 & \better +2.6 & \muchworse -7.5 & -0.1 & \muchworse -7.7 & \muchworse -9.5 & \worse -1.6\\
\midrule
(B) & public & mBERT & \muchbetter +6.5 & +0.2 & \muchbetter +8.5 & \muchbetter +7.6 & \better +2.3 & \worse -4.4 & \muchbetter +6.9 & \muchworse -6.1 & \muchworse -8.4 & \worse -2.6 \\
(B) & public & XLM-R & +0.9 & \worse -2.9 & \muchbetter +9.5 & \muchbetter +9.0 & \worse -1.2 & \muchworse -10.0 & +0.0 & \muchworse -5.9 & \muchworse -8.8 & \muchworse -6.3\\
\midrule
(C) & public & mBERT$_\textit{ft}$ & \muchbetter +7.8 & \muchbetter \textbf{+5.6} & \muchbetter +7.7 & \muchbetter +10.0 & \better +4.1 & -0.6 & \muchbetter +7.4 & \muchworse -8.0 & \muchworse -6.8 & +0.3\\
(C) & public & XLM-R$_\textit{ft}$ & \muchbetter +7.7 & \better +4.9 & \muchbetter +6.2 & \muchbetter +9.3 & \better +4.5 & \worse -2.6 & \muchbetter +7.0 & \muchworse -9.0 & \muchworse -7.6 & \better +1.0 \\
(C) & public & XLM-R$_\textit{ft.s}$ & \muchbetter +7.3 & \better +1.5 & \muchbetter +10.1 & \muchbetter \textbf{+12.4} & \better +4.8 & \worse -3.0 & \muchbetter +9.1 & \worse -3.8 & \worse -3.7 & \best \textbf{+2.3}\\
\midrule
(D) & public & GBv4$_\textit{ft}$ & \muchbetter +8.5 & \better +4.3 & \muchbetter +5.9 & \muchbetter +8.9 & \muchbetter +5.0 & \worse -1.5 & \muchbetter +7.7 & \muchworse -9.4 & \muchworse -9.1 & -0.1\\
(D) & public & L128K$_\textit{ft}$ & \muchbetter +6.4 & \better +3.1 & \muchbetter +6.5 & \muchbetter +8.2 & \better +1.6 & \worse -1.6 & \muchbetter +6.1 & \muchworse -9.0 & \muchworse -9.4 & \worse -3.6\\
(D) & public & L128K$_\textit{ft.s}$ & \muchbetter +7.0 & \better +3.7 & \muchbetter \textbf{+10.3} & \muchbetter +11.8 & \muchbetter \textbf{+5.4} & -0.3 & \muchbetter +5.2 & \worse -4.4 & \worse -4.6 & \better +2.1 \\
\midrule
(E) & GBv4 & mBERT$_\textit{ft}$ & \muchbetter +8.4 & \better +3.2 & \muchbetter +7.7 & \muchbetter +9.9 & \better +4.7 & \worse -1.5 & \better +3.2 & \muchworse -7.1 & \muchworse -6.7 & +0.7\\
(E) & GBv4 & XLM-R$_\textit{ft}$ & \muchbetter +9.6 & \better +1.8 & \muchbetter +7.0 & \muchbetter +9.5 & \muchbetter +5.2 & -0.4 & \better +1.4 & \muchworse -8.3 & \muchworse -7.7 & \better +1.4\\
(E) & L128K & mBERT$_\textit{ft}$ & \muchbetter \textbf{+12.1} & \better +3.3 & \muchbetter +7.9 & \muchbetter +9.9 & \better +4.7 & \worse -1.4 & \muchbetter +7.2 & \muchworse -8.1 & \muchworse -6.7 & \better +1.3\\
(E) & L128K & XLM-R$_\textit{ft}$ & \muchbetter +10.2 & \worse -1.9 & \muchbetter +6.1 & \muchbetter +9.4 & \better +4.8 & -0.5 & \better +4.6 & \muchworse -9.8 & \muchworse -7.5 & \better +2.0\\
\midrule
(S) & public & ST & - & \muchbetter +5.5 & +0.1 & \muchworse -20.3 & +0.3 & - & \best \textbf{+10.0} & \best \textbf{+1.8} & \muchworse -29.6 & \better +1.2 \\
\midrule
\midrule
& & &\multicolumn{5}{l||}{\textit{GBv4 (base, bilingual)}} & \multicolumn{5}{l}{\textit{L128K (large, bilingual)}}  \\
\midrule
(Z) & - & - & 46.0 & 45.4 & 64.7 & 33.2 & 41.7 & 42.7 & 46.3 & 67.9 & 36.7 & 40.9\\
\midrule
(C) & public & mBERT$_\textit{ft}$ & +0.6 & \better +3.7 & \better +2.6 & \muchbetter +6.9 & \muchbetter +7.5 & \better +2.7 & \muchbetter +8.2 & -0.9 & \better +4.9 & \muchbetter +11.7\\
(C) & public & XLM-R$_\textit{ft}$ & \worse -1.4 & \better \textbf{+4.5} & \better +1.8 & \muchbetter +6.0  & \muchbetter +8.4 & \better +1.2 & \muchbetter \textbf{+9.0} & \worse -2.5 & \better +3.9 & \muchbetter +10.5 \\
(C) & public & XLM-R$_\textit{ft.s}$ & -0.1 & \better +3.4 & \muchbetter \textbf{+5.1} & \muchbetter \textbf{+9.2} & \muchbetter +8.0 & \better +2.7 & \muchbetter +7.0 & \better +1.2 & \muchbetter \textbf{+7.2} & \muchbetter \textbf{+12.1}\\
\midrule
(E) & GBv4 & mBERT$_\textit{ft}$ & -0.1 & +0.1 & \better +3.3 & \muchbetter +7.2 & \muchbetter +8.1 & \better +4.2 & -0.5 & -0.1 & \muchbetter +5.1  & \muchbetter +11.2\\
(E) & GBv4 & XLM-R$_\textit{ft}$ & +0.1 & +0.4 & \better +1.5 & \muchbetter +6.0  & \muchbetter \textbf{+9.7} & \better +2.4 & +0.0 & \worse -1.3 & \better +4.2  & \muchbetter +10.8\\
(E) & L128K & mBERT$_\textit{ft}$ & -0.6 & \better +1.0 & \better +2.6 & \muchbetter +6.1  &  \muchbetter +7.4& \best \textbf{+5.5} & +0.8 & -0.7 & \better +4.7  & \muchbetter +10.6 \\
(E) & L128K & XLM-R$_\textit{ft}$ & +0.9 & \worse -2.1 & \better +1.1 & \muchbetter +5.5  & \muchbetter +7.8 & \better +4.4 & \worse -3.6 & \worse -2.2 & \better +4.1  & \muchbetter +11.3\\
\midrule
(F) & GBv4 & GBv4$_\textit{ft}$ & +0.0 & \worse -1.9 & \better +1.6 & \better +4.5 & \muchbetter +9.1  & \better +2.0 & -0.3 & \worse -1.7 & \better +3.2  & \muchbetter +10.9\\
(F) & GBv4 & L128K$_\textit{ft}$ & -0.9 & \worse -1.4 & \better +1.5 & \better +4.1 & \muchbetter +5.7  & \better +2.3 & \worse -1.7 & \worse -2.4 & \better +2.6 & \muchbetter +8.3\\
(F) & L128K & GBv4$_\textit{ft}$ & \worse -4.3 & \worse -1.0 & +0.4 & \better +4.1 & \muchbetter +7.4 & \better +4.1 & \worse -3.6 & \worse -2.1 & \better +2.3 & \muchbetter +11.4\\
(F) & L128K & L128K$_\textit{ft}$ & \worse -3.5 & \worse -1.1 & +0.3 & \better +3.8  & \better + 4.5 & \better +2.9 & +0.1 & \worse -2.9 & \better +2.0 & \muchbetter +6.7\\
(F) & L128K & L128K$_\textit{ft.s}$ & \better \textbf{+1.9} & +0.2 & \better +3.3 & \muchbetter +7.4 & \muchbetter +7.2 & \better +2.8 & \worse -1.8 & +0.8 & \muchbetter \muchbetter +6.0 & \muchbetter +11.8 \\
\midrule
(S) & public & ST & - & \worse -2.5 & \worse -1.3 & \muchworse -18.6 & \better +1.9 & - & \muchbetter +7.1 & \better \textbf{+1.5} & \muchworse -21.7 & \muchbetter +8.1\\
\bottomrule
\end{tabular}
\caption{Performance of Arabic on 5 tasks under various setups. %
Cells are colored by performance difference over zero-shot baseline: \colorbox{muchbetter}{+5 or more}, \colorbox{better}{+1 to +5}, \colorbox{worse}{-1 to -5}, \colorbox{muchworse}{-5 or more}. \colorbox{best}{\textbf{Highlights}} indicate the best setting for each task (best viewed in color). The best setting for each task and encoder combination is \textbf{bolded}. We order four encoders along two axes, similar to \autoref{tab:encoder}. 
}
\vspace{-8mm}
\label{tab:main}
\end{center}
\end{table*}

In \autoref{tab:main}, we present the Arabic test performance of five tasks under all  combinations considered. The ``MT'' and ``Align'' columns indicate the models used for the translation and word alignment components of the data projection process. For ACE, we report results on the average of six metrics.\footnote{Six metrics include entity, relation, trigger identification and classification, and argument identification and classification accuracies. See \autoref{app:ace-ar-full} for a breakdown of metrics.} For a large bilingual encoder, we use L128K instead of L64K due to its slightly better performance on English ACE (\autoref{app:ace-en-full}).

\paragraph{Impact of Data Projection} By comparing any group against group Z, we observe adding silver data yields better or equal performance to zero-shot in at least some setup in the IE tasks (ACE, NER, and BETTER). For syntax-related tasks, we observe similar trends, with the exception of XLM-R. We hypothesize that XLM-R provides better syntactic cues than those obtainable from the alignment, which we discuss later in relation to self-training.

\paragraph{Impact of Word Aligner} By comparing groups A, B, and C of the same encoder, we observe that \awesome~performs overall better than statistical MT-based fast-align (FA). Additional fine-tuning ($\textit{ft}$) on MT training bitext further improves its performance. As a result, we use fine-tuned aligners for further experiments. Moreover, incorporating supervised signals from gold alignments in the word alignment component ($\textit{ft.s}$) often helps performance of the task. In terms of computation budget, these three groups use a publicly available MT system \citep[``public'';][]{tiedemann-2020-tatoeba} and require only fine-tuning the encoder for alignment, which requires small additional computation.%

\paragraph{Impact of Encoder Size} Large bilingual or multilingual encoders tend to perform better than base encoders in the zero-shot scenario, with the exception of the bilingual encoders on ACE and BETTER.
While we observe base size encoders benefit from reducing the number of supported languages (from 100 to 2), for large size encoders trained much longer, the zero-shot performance of the bilingual model is worse than that of the multilingual model.
After adding silver data from group C based on the public MT model and the fine-tuned aligner, the performance gap between base and large models tends to shrink, with the exception of both bilingual and multilingual encoders on NER. %
In terms of computation budget, training a bilingual encoder requires significant additional computation.

\paragraph{Impact of Encoder on Word Aligner} By comparing groups C and D (in multilingual encoders) or groups E and F (in bilingual encoders), we observe bilingual encoders tend to perform slightly worse than multilingual encoders for word alignment. If bilingual encoders exist, using them in aligners requires little additional computation.

\paragraph{Impact of Encoder on MT} By comparing groups C and E, we observe the performance difference between the bilingual encoder based MT and the public MT depends on the task and encoder, and neither MT system clearly outperforms the other in all settings, despite the bilingual encoder having a better BLEU score.
The results suggest that both options should be explored if one's budget allows.
In terms of computation budget, using pretrained encoders in a custom MT system requires medium additional computation.%

\begin{table*}[th]
\small
\begin{center}
\begin{tabular}{ll ccccccccc c}

\toprule

\textbf{Encoder} & \textbf{Data} & \textbf{ar} & \textbf{de} & \textbf{en} & \textbf{es} & \textbf{fr} & \textbf{hi} & \textbf{ru} & \textbf{vi} & \textbf{zh} & \textbf{Average} \\

\midrule
\multicolumn{12}{l}{\textit{NER (F1)}} \\ 
\midrule
mBERT & Zero-shot & 41.6 & \textbf{78.8} & 83.9 & 73.1 & 79.5 & 66.2 & 63.4 & 70.8 & 51.8 & 67.7 \\
 & + Self & \muchbetter \textbf{+7.7} & -0.5 & \textbf{+0.4} & \better +4.8 & \best \textbf{+2.4} & \worse -2.5 & \better \textbf{+2.7} & \better \textbf{+1.2} & \better +1.4 & \better \textbf{+2.0} \\
 & + Proj & \muchworse -5.8 & -0.6 & +0.3 & \better +3.6 & +0.2 & \textbf{+0.4} & \worse -1.7 & \worse -2.0 & \better \textbf{+2.3} & -0.4 \\
 & + Proj (Bi) & +0.3 & -0.7 & +0.1 & \best \textbf{+5.2} & -0.6 & \worse -2.1 & \worse -1.1 & +0.3 & +0.0 & +0.2 \\
\midrule
XLM-R & Zero-shot & 46.4 & 79.5 & 83.9 & 76.1 & 80.0 & 70.9 & \best \textbf{70.5} & 77.0 & 40.2 & 69.4 \\
 & + Self & \best \textbf{+11.2} & \best \textbf{+0.9} & \best \textbf{+0.6} & \better \textbf{+1.0} & \textbf{+0.5} & \better +2.1 & \worse -1.5 & \best \textbf{+1.7} & \better +2.3 & \best \textbf{+2.1} \\
 & + Proj & \better +1.7 & -0.7 & -0.1 & \worse -3.9 & \worse -1.2 & \better +1.2 & \worse -4.8 & \muchworse -9.1 & \muchbetter +14.2 & -0.3 \\
 & + Proj (Bi) & \muchbetter +6.9 & +0.4 & -0.2 & \worse -4.3 & \worse -1.5 & \best \textbf{+3.2} & \worse -3.3 & \muchworse -5.2 & \best \textbf{+15.1} & \better +1.2 \\
\midrule
\multicolumn{12}{l}{\textit{POS (ACC)}} \\ 
\midrule
mBERT & Zero-shot & 59.7 & 89.6 & \textbf{96.9} & 87.5 & 88.7 & 69.5 & 81.9 & 62.6 & 66.6 & 78.1 \\
 & + Self & +0.3 & \textbf{+0.5} & +0.0 & \textbf{+0.4} & \textbf{+0.4} & -0.3 & \textbf{+0.5} & \textbf{+0.4} & \best \textbf{+1.7} & \textbf{+0.4} \\
 & + Proj & \muchbetter +6.9 & \worse -3.2 & +0.0 & \worse -3.8 & \worse -3.9 & \better +1.3 & \muchworse -6.6 & \muchworse -7.4 & \worse -4.1 & \worse -2.3 \\
 & + Proj (Bi) & \muchbetter \textbf{+8.5} & \worse -2.6 & -0.1 & \worse -3.2 & \worse -3.0 & \better \textbf{+1.6} & \muchworse -5.7 & \muchworse -6.9 & \worse -3.9 & \worse -1.7 \\
\midrule
XLM-R & Zero-shot & 73.3 & \best \textbf{91.5} & \best \textbf{98.0} & \best \textbf{89.3} & \best \textbf{90.0} & 78.6 & 86.8 & \best \textbf{65.2} & 53.6 & 80.7 \\
 & + Self & \best \textbf{+1.6} & -0.3 & +0.0 & +0.0 & +0.0 & \better \best \textbf{+2.0} & \best \textbf{+0.1} & -0.4 & \muchbetter \textbf{+11.7} & \best \textbf{+1.6} \\
 & + Proj & \muchworse -7.1 & \muchworse -5.4 & -0.5 & \muchworse -6.3 & \muchworse -5.9 & \muchworse -6.0 & \muchworse -10.5 & \muchworse -8.9 & \muchbetter +9.7 & \worse -4.6 \\
 & + Proj (Bi) & \muchworse -6.1 & \worse -4.6 & -0.1 & \worse -4.9 & \worse -4.6 & \muchworse -5.5 & \muchworse -10.4 & \muchworse -8.7 & \muchbetter +9.4 & \worse -4.0 \\
\midrule
\multicolumn{12}{l}{\textit{Parsing (LAS)}} \\ 
\midrule
mBERT & Zero-shot & 29.2 & \textbf{67.7} & 79.7 & \textbf{68.9} & \textbf{73.2} & 31.2 & \textbf{60.6} & \textbf{33.6} & \textbf{29.4} & \textbf{52.6} \\
 & + Self & \muchworse -20.6 & \muchworse -34.2 & +0.1 & \muchworse -41.6 & \muchworse -41.1 & \muchworse -15.3 & \muchworse -35.2 & \muchworse -17.8 & \muchworse -14.5 & \muchworse -24.5 \\
 & + Proj & \muchbetter \textbf{+9.1} & \worse -2.1 & \better \textbf{+1.1} & \worse -4.9 & \muchworse -5.8 & \muchbetter \textbf{+6.0} & \muchworse -5.6 & \muchworse -7.2 & \worse -2.1 & \worse -1.3 \\
 & + Proj (Bi) & \muchbetter +7.6 & \worse -1.6 & +0.5 & \worse -3.8 & \worse -4.5 & \muchbetter +5.7 & \worse -4.8 & \muchworse -7.2 & \worse -2.5 & \worse -1.2 \\
\midrule
XLM-R & Zero-shot & \best \textbf{48.0} & \best \textbf{69.6} & 82.6 & \best \textbf{73.6} & \best \textbf{76.1} & \best \textbf{43.1} & \best \textbf{70.3} & \best \textbf{38.4} & 15.0 & \best \textbf{57.4} \\
 & + Self & \muchworse -30.4 & \muchworse -29.4 & \best \textbf{+0.1} & \muchworse -39.9 & \muchworse -40.0 & \muchworse -18.3 & \muchworse -33.9 & \muchworse -16.1 & \muchworse -9.7 & \muchworse -24.2 \\
 & + Proj & \muchworse -8.5 & \worse -4.3 & +0.0 & \muchworse -10.3 & \muchworse -10.1 & \muchworse -5.7 & \muchworse -14.8 & \muchworse -11.1 & \muchbetter +14.5 & \muchworse -5.6 \\
 & + Proj (Bi) & \muchworse -8.4 & \worse -1.6 & \best \textbf{+0.1} & \muchworse -7.7 & \muchworse -7.4 & \worse -3.1 & \muchworse -12.7 & \muchworse -9.8 & \best \textbf{+15.1} & \worse -3.9 \\
\bottomrule
\end{tabular}
\caption{Performance of NER, POS, and parsing for eight target languages. We use the same color code as \autoref{tab:main}.}
\label{tab:multi}
\end{center}
\end{table*}

\paragraph{Impact of Label Source} 
To assess the quality of the projected annotations in the silver data, we consider a different way to automatically label translated sentences: self-training \citep[ST;][]{yarowsky-1995-unsupervised}. For self-training, we translate the source data to the target language, label the translated data using a zero-shot model trained on source data, and combine the labeled translations with the source data to train a new model.\footnote{This setup differs from traditional zero-shot self-training in cross-lingual transfer, as the traditional setup assumes unlabeled corpora in the target language(s) \cite{eisenschlos-etal-2019-multifit} instead of translations of the source language data.}
Compared to the silver data, the self-training data has the same underlying text but a different label source. 

We first observe that self-training for parsing leads to significantly worse performance due to the low quality of the predicted trees. 
By comparing groups S and C, which use the same underlying text, we observe that data projection tends to perform better than self-training, with the exceptions of POS tagging with a large encoder and NER with a large multilingual encoder. 
These results suggest that the external knowledge\footnote{``External knowledge'' refers to knowledge introduced into the downstream model as a consequence of the particular decisions made by the aligner (and subsequent projection).}
in the silver data complements the knowledge obtainable when the model is trained with source language data alone, but when the zero-shot model is already quite good (like for POS tagging) data projection can harm performance compared to self-training.
Future directions could include developing task-specific projection and alignment heuristics to improve projected annotation quality for POS tagging or parsing, and combining data projection and self-training.

\subsection{Multilingual Experiments}

In \autoref{tab:multi}, we present the test performance of three tasks for eight target languages. We use the public MT system \cite{tiedemann-2020-tatoeba} and non-fine-tuned \awesome~with mBERT as the word aligner for data projection---a setup with the smallest computation budget---due to computation constraints. We consider both data projection (+Proj) and self training (+Self). We use silver data in addition to English gold data for training. We use multilingual training with +Self and +Proj, and bilingual training with +Proj (Bi).

We observe that data projection (+Proj (Bi)) sometimes benefits languages with the lowest zero-shot performance (Arabic, Hindi, and Chinese), with the notable exception of XLM-R on syntax-based tasks (excluding Chinese). For languages closely related to English, data projection tends to hurt performance. %
We observe that for data projection, training multiple bilingual models (+Proj (Bi)) outperforms joint multilingual training (+Proj). This could be the result of noise from alignments of various quality mutually interfering. In fact, self-training with the same translated text (+Self) outperforms data projection and zero-shot scenarios, again with the exception of parsing. As data projection and self-training use the same translated text and differ only by label source, the results indicate that the external knowledge from frozen mBERT-based alignment is worse than what the model learns from source language data alone. Thus, further performance improvement could be achieved with an improved aligner.

\section{Related Work}

Although projected data may be of lower quality than the original source data due to errors in translation or alignment, it is useful for tasks such as semantic role labeling \cite{akbik-etal-2015-generating,aminian-etal-2019-cross}, information extraction \cite{riloff-etal-2002-inducing}, POS tagging \cite{yarowsky-ngai-2001-inducing}, and dependency parsing \cite{ozaki-etal-2021-project}. The intuition is that although the projected data may be noisy, training on it gives a model useful information about the statistics of the target language.

\citet{akbik-etal-2015-generating} and \citet{aminian-etal-2017-transferring} use bootstrapping algorithms to iteratively construct projected datasets for semantic role labeling. \citet{akbik-etal-2015-generating} additionally use manually defined filters to maintain high data quality, which results in a projected dataset that has low recall with respect to the source corpus.
\citet{fei-etal-2020-cross} and \citet{daza-frank-2020-x} find that a non-bootstrapped approach works well for cross-lingual SRL. Advances in translation and alignment quality allow us to avoid bootstrapping while still constructing projected data that is useful for downstream tasks.

\citet{fei-etal-2020-cross} and \citet{daza-frank-2020-x} also find improvements when training on a mixture of gold source language data and projected silver target language data.
Ideas from domain adaptation can be used to make more effective use of gold and silver data to mitigate the effects of language shift \cite{xu-etal-2021-gradual}.

Improvements to task-specific models for zero-shot transfer are orthogonal to our work. For example, language-specific information can be incorporated using language indicators or embeddings \cite{johnson-etal-2017-googles}, contextual parameter generators \cite{platanios-etal-2018-contextual}, or language-specific semantic spaces \cite{luo2021revisiting}. Conversely, adversarial training \cite{ganin2016domain} has been used to discourage models from learning language-specific information \cite{chen-etal-2018-adversarial,keung-etal-2019-adversarial,ahmad-etal-2019-cross}.

\section{Conclusion}

In this paper, we explore data projection and the use of silver data in zero-shot cross-lingual IE, facilitated by neural machine translation and word alignment. Recent advances in pretrained encoders have improved machine translation systems and word aligners in terms of intrinsic evaluation. We conduct an extensive extrinsic evaluation and study how the encoders themselves---and components containing them---impact performance on a range of downstream tasks and languages.

With a test bed of English--Arabic IE tasks, we find that adding projected silver training data overall yields improvements over zero-shot learning. Comparisons of how each factor in the data projection process impacts performance show that while one might hope for the existence of a silver bullet strategy, the best setup is usually task dependent.

In multilingual experiments, we find that silver data tends to help languages with the weakest zero-shot performance, and that it is best used separately for each desired language pair instead of in joint multilingual training.

We also examine self-training with translated text to assess when data projection helps cross-lingual transfer, and find it to be another viable option for obtaining labels for some tasks.
In future work, we will explore how to improve alignment quality and how to combine data projection and self-training techniques. 

\section*{Acknowledgments}
We thank the anonymous reviewers for their valuable comments. We thank João Sedoc for helpful discussions and Shabnam Behzad for post-submission experiments. This work was supported in part by IARPA
BETTER (\#2019-19051600005). The views and conclusions contained in this work are those of the authors and should not be interpreted as necessarily representing the official policies, either expressed or implied, or endorsements of ODNI, IARPA, or the U.S. Government. The U.S. Government is authorized to reproduce and distribute reprints for governmental purposes notwithstanding any copyright annotation therein.

This research made use of the following open-source software: AllenNLP \citep{gardner-etal-2018-allennlp}, FairSeq \citep{ott-etal-2019-fairseq}, NumPy \citep{2020NumPy-Array}, PyTorch \citep{paszke2017automatic}, PyTorch lightning \cite{falcon2019pytorch}, scikit-learn \citep{JMLR:v12:pedregosa11a}, and Transformers \citep{wolf-etal-2020-transformers}.

\bibliographystyle{acl_natbib}
\bibliography{anthology,acl2021}

\clearpage
\appendix

\section{Fine-tuning Hyperparameters}
\label{app:ft-hparams}
\subsection{ACE}
We used the OneIE v0.4.8 codebase\footnote{http://blender.cs.illinois.edu/software/oneie/} with the following hyperparameters: Adam optimizer \citep{kingma2014adam} for 60 epochs with a learning rate of 5e-5 and weight decay of 1e-5 for the encoder, and a learning rate of 1e-3 and weight decay of 1e-3 for other parameters. Two-layer feed-forward network with a dropout rate of 0.4 for task-specific classifiers, 150 hidden units for entity and relation extraction, and 600 hidden units for event extraction. $\beta_v$ and $\beta_e$ set to 2 and $\theta$ set to 10 for global features.
Data statistics can be found in \autoref{tab:ace-stats}.

\begin{table}[th]
\small
\begin{center}
\begin{tabular}{lccc}

\toprule
& Train (en/ar*) & Dev (en/ar*)  & Test (en/ar)\\
\midrule
Sent. &19,216/1,710&901/256&676/216\\
Evt. trig. & 4,419/1,825& 468/211&424/234 \\
Evt. arg. & 6,607/3,255& 759/412 & 689/451\\
Entity &47,554/25,889&3,423/3,554&3,673/2,977 \\
Relation &7,159/3,704&728/527& 802/478 \\
Rel. arg. & 14,318/7,408 & 1,456/1,054 & 1,604/956\\
\bottomrule

\end{tabular}
\caption{ACE dataset statistics. *Arabic train and dev sets are not used in our experiments.}
\label{tab:ace-stats}
\end{center}
\end{table}

\begin{table}[th]
\small
\begin{center}
\begin{tabular}{lccc}

\toprule
& Train (en) & Dev (en)  & Test (ar)\\
\midrule
Sent. &3,629& 453& 129\\
Evt. trig. & 12,390& 1,527& 517 \\
Evt. arg. & 20,314& 2,522& 857\\
\bottomrule

\end{tabular}
\caption{BETTER dataset statistics.}
\label{tab:better-stats}
\end{center}
\end{table}

\subsection{BETTER}
\label{app:better-span-finder}
The codebase for event structure prediction uses AllenNLP \citep{gardner-etal-2018-allennlp}.
The contextual encoder produces representations for the tagger and typer modules. 
Span representations are formed by concatenating the output of a self-attention layer over the span's token embeddings with the embeddings of the first and last tokens of the span. 
The BiLSTM-CRF tagger has 2 layers, both with hidden size of 2048.
We use a dropout rate of 0.3 and maximum sequence length of 512.
Child span prediction is conditioned on parent spans and labels, so we represent parent labels with an embedding of size 128. 
We use Adam optimizer to fine-tune the encoder with a learning rate of 2e-5, and we use a learning rate of 1e-3 for other components. 
The tagger loss is negative log likelihood and the typer loss is cross entropy.
We equally weight both losses and train against their sum. The contextual encoder is not frozen.
Data statistics can be found in \autoref{tab:better-stats}.

\subsection{NER, POS Tagging, and Parsing}
We use the Adam optimizer with a learning rate of 2e-5 with linear warmup for the first 10\% of total steps and linear decay afterwards, and train for 5 epochs with a batch size of 32.
To obtain valid BIO sequences, we rewrite standalone \texttt{I-X} into \texttt{B-X} and \texttt{B-X I-Y I-Z} into \texttt{B-Z I-Z I-Z}, following the final entity type. For parsing, we ignore punctuations (PUNCT) and symbols (SYM) when calculating LAS.

We set the maximum sequence length to 128 during fine-tuning. For NER and POS tagging, we additionally use a sliding window of context to include subwords beyond the first 128. At test time, we use the same maximum sequence length except for parsing. At test time for parsing, we use only the first 128 words of a sentence.  As the supervision for Chinese NER is character-level, we segment the characters into words using the Stanford Word Segmenter and realign the label.

The datasets we used are publicly available: NER,\footnote{\url{https://www.amazon.com/clouddrive/share/d3KGCRCIYwhKJF0H3eWA26hjg2ZCRhjpEQtDL70FSBN}} POS tagging, and dependency parsing.\footnote{\url{https://lindat.mff.cuni.cz/repository/xmlui/handle/11234/1-3424}}
Data statistics can be found in \autoref{tab:other-stat}.

\begin{table}[th]
\small
\begin{center}
\begin{tabular}[b]{ccc}
\toprule
  & \multirow{2}{*}{NER} & POS tagging \\
  & & Parsing\\
\midrule
en-train & 20,000 & 12,543 \\
en-dev & 10,000 & 2,002 \\
en-test & 10,000 & 2,077 \\
\midrule
ar-test & 10,000 & 680 \\
de-test & 10,000 & 977 \\
es-test & 10,000 & 426 \\
fr-test & 10,000 & 416 \\
hi-test & 1,000 & 1,684 \\
ru-test & 10,000 & 601 \\
vi-test & 10,000 & 800 \\
zh-test & 10,000 & 500 \\
\bottomrule
\end{tabular}
\caption{Number of examples.\label{tab:other-stat}}
\end{center}
\end{table}

\section{Encoder Pretraining Hyperparameters}
\label{app:pretrain-hparams}
We pretrain each encoder with a batch size of 2048 sequences and 512 sequence length for 250K steps from scratch,\footnote{While we use XLM-R as the initialization of the Transformer, due to vocabulary differences, the learning curve is similar to that of pretraining from scratch.} roughly 1/24 the amount of pretraining compute of XLM-R. Training takes 8 RTX 6000 GPUs roughly three weeks. We follow the pretraining recipe of RoBERTa \cite{liu2019roberta} and XLM-R. We omit the next sentence prediction task and use a learning rate of 2e-4, Adam optimizer, and linear warmup of 10K steps then decay linearly to 0, multilingual sampling alpha of 0.3, and the fairseq \cite{ott-etal-2019-fairseq} implementation.

\section{Machine Translation Hyperparameters}
\label{app:mt-hparams}
All of our machine translation systems are based upon the Transformer architecture: a six-layer encoder, six-layer decoder model with 2048 FFN dimension and 8 attention heads. We use 4 Nvidia V100 GPUs, with a batch size of 2048 tokens per GPU. We accumulate the gradient 10 times before updating model parameters. The initial learning rate is 1e-3. The optimizer is Adam with an \texttt{inverse\_sqrt} learning rate scheduler. In the inference step, the width of beam search is 4 with a length penalty of 0.6.

\section{ACE Arabic Full Metrics}
\label{app:ace-ar-full}

The full metrics of Arabic ACE can be found in \autoref{tab:ace-full-1} and \autoref{tab:ace-full-2}. 

\begin{table*}[]
\small
\begin{center}
\begin{tabular}{lll cccccc c}

\toprule
 & \textbf{MT} & \textbf{Align} & \textbf{Entity} & \textbf{Relation} & \textbf{Trig-I} & \textbf{Trig-C} & \textbf{Arg-I} & \textbf{Arg-C} & \textbf{AVG}\\
\midrule
\multicolumn{10}{l}{\textit{mBERT (base, multilingual)}} \\
\midrule
(Z) & - & - & 59.3 & 25.7 & 23.8 & 22.2 & 17.2 & 13.8 & 27.0 \\
\midrule
(A) & public & FA & -2.2 & -13.9 & +6.5 & +2.5 & +10.7 & +11.5 & \better +2.5 \\
\midrule
(B) & public & mBERT & -6.2 & -5.1 & +16.0 & +10.6 & +11.5 & +12.1 & \muchbetter +6.5 \\
(B) & public & XLM-R & -12.7 & -17.9 & +11.1 & +8.0 & +8.5 & +8.1 & +0.9 \\
\midrule
(C) & public & mBERT$_\textit{ft}$ & -1.1 & +0.9 & +12.8 & +9.8 & +10.9 & +13.6 & \muchbetter +7.8 \\
(C) & public & XLM-R$_\textit{ft}$ & -0.1 & -4.2 & +16.0 & +11.9 & +11.2 & +11.3 & \muchbetter +7.7 \\
(C) & public & XLM-R$_\textit{ft.s}$ & -0.2 & -1.6 & +13.4 & +11.5 &+9.0& +11.7 & \muchbetter +7.3\\
\midrule
(D) & public & GBv4$_\textit{ft}$ & -1.9 & +2.8 & +14.3 & +9.9 & +12.7 & +13.3 & \muchbetter +8.5 \\
(D) & public & L128K$_\textit{ft}$ & -1.7 & +0.6 & +11.6 & +8.3 & +10.7 & +9.0 & \muchbetter +6.4 \\
(D) & public & L128K$_\textit{ft.s}$ & -1.3 & +3.6& +12.7& +8.4& +8.3& +10.3 & \muchbetter +7.0\\
\midrule
(E) & GBv4 & mBERT$_\textit{ft}$ & +1.0 & +4.7 & +13.6 & +10.3 & +9.3 & +11.3 & \muchbetter +8.4 \\
(E) & GBv4 & XLM-R$_\textit{ft}$ & -0.5 & +5.5 & +12.6 & +10.8 & +15.1 & +14.4 & \muchbetter +9.6 \\
(E) & L128K & mBERT$_\textit{ft}$ & +2.6 & +5.2 & +12.9 & +13.4 & +18.8 & +19.6 & \muchbetter +12.1 \\
(E) & L128K & XLM-R$_\textit{ft}$ & +2.5 & +6.3 & +11.2 & +5.1 & +17.1 & +19.2 & \muchbetter +10.2 \\
\midrule
\multicolumn{10}{l}{\textit{XLM-R (large, multilingual)}} \\
\midrule
(Z) & - & - & 70.0 & 38.7 & 44.0 & 40.8 & 39.5 & 37.8 & 45.1 \\
\midrule
(A) & public & FA & -7.2 & -9.5 & -9.3 & -8.2 & -4.8 & -6.0 & \muchworse -7.5 \\
\midrule
(B) & public & mBERT & -8.5 & -10.2 & -2.2 & -0.1 & -2.0 & -3.4 & \worse -4.4 \\
(B) & public & XLM-R & -14.7 & -12.1 & -8.9 & -7.8 & -8.1 & -8.3 & \muchworse -10.0 \\
\midrule
(C) & public & mBERT$_\textit{ft}$ & -2.5 & +3.5 & -2.3 & -4.3 & +1.8 & +0.2 & -0.6 \\
(C) & public & XLM-R$_\textit{ft}$ & -3.8 & -0.5 & -3.6 & -4.5 & -0.5 & -2.8 & \worse -2.6 \\
(C) & public & XLM-R$_\textit{ft.s}$ & -2.4 & +1 & -3.6 & -7.5& -2.1& -3.2 & \worse -3.0\\
\midrule
(D) & public & GBv4$_\textit{ft}$ & -4.4 & +0.8 & +0.8 & -2.3 & -1.1 & -2.8 & \worse -1.5 \\
(D) & public & L128K$_\textit{ft}$ & -2.1 & -1.4 & +0.6 & -2.2 & -2.0 & -3.0 & \worse -1.6 \\
(D) & public & L128K$_\textit{ft.s}$ & -0.9 & +2.2& -1.6 & -4.4& +1.8 & +0.8 & -0.3\\
\midrule
(E) & GBv4 & mBERT$_\textit{ft}$ & -2.5 & +1.4 & -3.2 & -4.1 & +0.3 & -0.9 & \worse -1.5 \\
(E) & GBv4 & XLM-R$_\textit{ft}$ & -2.2 & +2.3 & -2.2 & -3.0 & +2.9 & -0.3 & -0.4 \\
(E) & L128K & mBERT$_\textit{ft}$ & +0.1 & -1.1 & -2.2 & -3.0 & -0.9 & -1.3 & \worse -1.4 \\
(E) & L128K & XLM-R$_\textit{ft}$ & +0.1 & +4.2 & -4.4 & -6.3 & +1.8 & +1.3 & -0.5 \\
\midrule
\multicolumn{10}{l}{\textit{GBv4 (base, bilingual)}} \\
\midrule
(Z) & - & - & 71.9 & 29.6 & 49.8 & 46.8 & 41.1 & 36.8 & 46.0 \\
\midrule
(C) & public & mBERT$_\textit{ft}$ & -0.1 & +9.3 & -5.8 & -5.8 & +3.0 & +3.0 & +0.6 \\
(C) & public & XLM-R$_\textit{ft}$ & -0.8 & +10.4 & -8.0 & -9.0 & -1.5 & +0.6 & \worse -1.4 \\
(C) & public & XLM-R$_\textit{ft.s}$ & +0.0 & +9.7 & -8.0 & -7.0 & +1.7 & +2.9 &-0.1\\
\midrule
(E) & GBv4 & mBERT\_FT & -2.0 & +7.6 & -4.9 & -5.8 & +1.8 & +2.5 & -0.1 \\
(E) & GBv4 & XLMR\_FT & -1.1 & +9.8 & -8.5 & -7.7 & +3.8 & +4.2 & +0.1 \\
(E) & L128K & mBERT\_FT & -3.0 & +8.1 & -4.9 & -3.9 & -0.8 & +0.9 & -0.6 \\
(E) & L128K & XLMR\_FT & -0.4 & +11.6 & -5.8 & -4.9 & +1.4 & +3.4 & +0.9 \\
\midrule
(F) & GBv4 & GBv4$_\textit{ft}$ & -1.9 & +8.1 & -4.7 & -4.6 & +1.7 & +1.7 & +0.0 \\
(F) & GBv4 & L128K$_\textit{ft}$ & -0.5 & +5.4 & -5.1 & -5.2 & -0.6 & +0.6 & -0.9 \\
(F) & L128K & GBv4$_\textit{ft}$ & -0.6 & +7.6 & -4.5 & -5.4 & -0.2 & +0.3 & \worse -4.3 \\
(F) & L128K & L128K$_\textit{ft}$ & -1.1 & +8.6 & -6.6 & -4.7 & +4.2 & +5.8 & \worse -3.5 \\
(F) & L128K & L128K$_\textit{ft.s}$ & +0.0 & +10.5 & -4.6& -5.6& +4.9& +6.0 & \better +1.9\\
\bottomrule
\end{tabular}
\caption{Detailed performance of bilingual English--Arabic ACE. Cells are colored following \autoref{tab:main}.}
\label{tab:ace-full-1}
\end{center}
\end{table*}

\begin{table*}[]
\small
\begin{center}
\begin{tabular}{lll cccccc c}

\toprule
 & \textbf{MT} & \textbf{Align} & \textbf{Entity} & \textbf{Relation} & \textbf{Trig-I} & \textbf{Trig-C} & \textbf{Arg-I} & \textbf{Arg-C} & \textbf{AVG}\\

\midrule
\multicolumn{10}{l}{\textit{L128K (large, bilingual)}} \\
\midrule
(Z) & - & - & 66.0 & 30.7 & 44.0 & 43.0 & 37.4 & 35.4 & 42.7 \\
\midrule
(C) & public & mBERT$_\textit{ft}$ & +2.1 & +7.8 & +1.8 & -0.6 & +3.1 & +1.6 & \better +2.7 \\
(C) & public & XLM-R$_\textit{ft}$ & +2.6 & +5.2 & -1.7 & -4.2 & +3.2 & +1.9 & \better +1.2 \\
(C) & public & XLM-R$_\textit{ft.s}$ & +5.7 & +8.9 & -2.6 & -4.2 & +4.6 & +4.0 & \better +2.7 \\
\midrule
(E) & GBv4 & mBERT\_FT & +3.8 & +12.7 & +2.4 & -0.1 & +3.6 & +2.3 & \better +4.2 \\
(E) & GBv4 & XLMR\_FT & +2.6 & +11.9 & -2.6 & -5.1 & +3.7 & +3.4 & \better +2.4\\
(E) & L128K & mBERT\_FT & +3.8 & +8.3 & +2.2 & +0.6 & +8.9 & +9.0 & \muchbetter +5.5 \\
(E) & L128K & XLMR\_FT & +2.8 & +9.6 & +3.5 & +0.8 & +4.8 & +4.8 & \better +4.4 \\
\midrule
(F) & GBv4 & GBv4$_\textit{ft}$ & +1.9 & +6.7 & +1.3 & -1.5 & +2.0 & +1.6 & \better +2.0 \\
(F) & GBv4 & L128K$_\textit{ft}$ & +2.7 & +7.1 & -1.1 & -3.5 & +4.9 & +3.1 & \better +2.3 \\
(F) & L128K & GBv4$_\textit{ft}$ & +2.2 & +8.9 & +2.3 & -0.2 & +6.7 & +4.1 & \better +4.1 \\
(F) & L128K & L128K$_\textit{ft}$ & +3.5 & +7.0 & -0.3 & -2.1 & +4.9 & +3.9 & \better +2.9 \\
(F) & L128K & L128K$_\textit{ft.s}$ & +3.6 & +11.7& -1.1 & -3.7 & +3.3 & +2.9 & \better +2.8\\
\bottomrule
\end{tabular}
\caption{Detailed performance of bilingual English--Arabic ACE. Cells are colored following \autoref{tab:main}.}
\label{tab:ace-full-2}
\end{center}
\end{table*}

\section{ACE English Full Metrics}
\label{app:ace-en-full}

The full metrics of English ACE can be found in \autoref{tab:ace_en}.

\begin{table*}
\begin{small}
\centering
\begin{tabular}{l|c|c|ccccccc}

\toprule
\textbf{Model}& \textbf{Train}&\textbf{Test}&\textbf{Entity} & \textbf{Relation}  & \textbf{Trig-I} & \textbf{Trig-C} & \textbf{Arg-I} & \textbf{Arg-C} & \textbf{AVG}\\
\midrule
\citet{lin-etal-2020-joint}& en& en&89.6&	58.6&	75.6&	72.8&	57.3&	54.8 & 68.1\\
BERT$_{large}$&en& en& 90.2&	64.0	&75.7	&73.2&	59.5&	57.4 & 70.0\\
\midrule
mBERT &en& en&89.5	&56.7	&72.4	&69.2	&53.3	&50.5 & 65.3\\
GBv4 & en& en&90.2	&63.0&73.8&	71.4&	57.7&	55.4&68.6\\
XLM-R & en& en&90.9&		\textbf{64.4}&	75.3&	72.2&	58.4&	55.5 & 69.4\\
L64K& en& en&91.30&		64.0&	\textbf{75.45}&	73.0&	59.4&	57.4 & 70.1\\
L128K&en& en&\textbf{91.32} &	64.1&	75.39&	\textbf{73.5}&	\textbf{59.6}&	\textbf{57.7} & \textbf{70.3}\\
\bottomrule

\end{tabular}
\caption{ACE results with different encoders. All models are trained and tested on gold English data.}
\label{tab:ace_en}
\end{small}
\end{table*}

\end{document}